\documentclass[runningheads]{llncs}
\usepackage[T1]{fontenc}
\usepackage{graphicx}
\usepackage{booktabs}
\usepackage[table]{xcolor}
\usepackage{multirow}
\usepackage{pgfplots} 
\usepackage{subcaption}
\usetikzlibrary{patterns}
\usepackage{color,soul}
\pgfplotsset{compat=1.16} 
\usepackage{hyperref}
\usepackage{color}
\definecolor{lightgray}{gray}{0.75}
\begin{document}
\title{Ensembling Transformers for Cross-domain Automatic Term Extraction}

%
%

\author{Hanh Thi Hong Tran\inst{1,2,3}\orcidID{0000-0002-5993-1630}  \and 
Matej Martinc\inst{1}\orcidID{0000-0002-7384-8112} \and 
Andraz Pelicon\inst{1}\orcidID{0000-0002-2060-6670} \and 
Antoine Doucet\inst{3}\orcidID{0000-0001-6160-3356} \and 
Senja Pollak\inst{2}\orcidID{0000-0002-4380-0863}} 
\authorrunning{Tran et al.}

\institute{Jožef Stefan International Postgraduate School, \\ Jamova cesta 39, 1000 Ljubljana, Slovenia \\
\email{tran.hanh@ijs.si} \and
Jožef Stefan Institute, \\ Jamova cesta 39, 1000 Ljubljana, Slovenia \and
University of La Rochelle, \\ 23 Av. Albert Einstein, La Rochelle, France\\
}

\maketitle              

\fbox{\begin{minipage}{30.5em}
\noindent Please cite the paper as:\\
Hanh Thi Hong TRAN, Matej MARTINC, Andraz PELICON, Antoine DOUCET and Senja POLLAK (2022): Ensembling Transformers for Cross-domain Automatic Term Extraction
In: Tseng, YH., Katsurai, M., Nguyen, H.N. (eds.) From Born-Physical to Born-Virtual: Augmenting Intelligence in Digital Libraries. ICADL 2022. Lecture Notes in Computer Science, vol 13636, pp. pp. 90–100, 2022. Springer, Cham. \href{https://doi.org/10.1007/978-3-031-21756-2_7}{https://doi.org/10.1007/978-3-031-21756-2\_7}.
\end{minipage}}

\begin{abstract}
Automatic term extraction plays an essential role in domain language understanding and several natural language processing downstream tasks. In this paper, we propose a comparative study on the predictive power of Transformers-based pretrained language models toward term extraction in a multi-language cross-domain setting. Besides evaluating the ability of monolingual models to extract single- and multi-word terms, we also experiment with ensembles of mono- and multilingual models by conducting the intersection or union on the term output sets of different language models. Our experiments have been conducted on the ACTER corpus covering four specialized domains (Corruption, Wind energy, Equitation, and Heart failure) and three languages (English, French, and Dutch), and on the RSDO5 Slovenian corpus covering four additional domains (Biomechanics, Chemistry, Veterinary, and Linguistics). The results show that the strategy of employing monolingual models outperforms the state-of-the-art approaches from the related work leveraging multilingual models, regarding all the languages except Dutch and French if the term extraction task excludes the extraction of named entity terms. Furthermore, by combining the outputs of the two best performing models, we achieve significant improvements.

\end{abstract}

\newpage
\section{Introduction}
\label{sec:intro}

Automatic Term Extraction (ATE) is the task of identifying specialized terminology from the domain-specific corpora. By easing the time and effort needed to manually extract the terms, ATE is not only widely used for terminographical tasks (e.g., glossary construction \cite{maldonado2016self}, specialized dictionary creation \cite{le2010automating}, etc.) but it also contributes to several complex downstream tasks (e.g., machine translation \cite{wolf2011statistical}, information retrieval \cite{lingpeng2005improving}, sentiment analysis \cite{pavlopoulos2014aspect}, to cite a few).

With recent advances in natural language processing (NLP), a new family of deep neural approaches, namely Transformers \cite{vaswani2017attention}, has been pushing the state-of-the-art (SOTA) in several sequence-labeling semantic tasks, e.g., named entity recognition (NER) \cite{lample2016neural,hanh2021named} and machine translation \cite{yang2020towards}, among others. 
The TermEval 2020 Shared Task on Automatic Term Extraction, organized as part of the CompuTerm workshop \cite{rigouts2020termeval}, presented one of the first opportunities to systematically study and compare various ATE systems with the advent of The Annotated Corpora for Term Extraction Research (ACTER) dataset \cite{rigouts2020termeval,rigouts2020no}, a novel corpora covering four domains and three languages. Regarding Slovenian, the RSDO5\footnote{https://www.clarin.si/repository/xmlui/handle/11356/1470} corpus \cite{RSDOcorpus11} was created with texts from four specialized domains.
Inspired by the success of Transformers for ATE in the TermEval 2020, we propose an extensive study of their performance in a cross-domain sequence-labeling setting and evaluate different factors that influence extraction effectiveness. The experiments are conducted on two datasets: ACTER and RSDO5 corpora.

Our major contributions can be summarized as the three following points:
\begin{itemize}
    \item An empirical evaluation of several monolingual and multilingual Transformer-based language models, including both masked (e.g., BERT and its variants) and autoregressive (e.g., XLNet) models, on the cross-domain ATE tasks;
    \item Filling the research gap in ATE task for Slovenian by experimenting with different models to achieve a new SOTA in the RSDO5 corpus. 
    \item An ensembling Transformer-based model for ATE that further improves the SOTA in the field.
\end{itemize}


This paper is organized as follows: Section \ref{sec:sota} presents the related work in term extraction. Next, we introduce our methodology in Section \ref{sec:method}, including the dataset description, the workflow, and experimental settings, as well as the evaluation metrics. The corresponding results are presented in Section \ref{sec:results}. Finally, we conclude the paper and present future directions in Section \ref{sec:conclusion}.
\section{Related work}
\label{sec:sota}

The research into monolingual ATE was first introduced during the 1990s \cite{damerau1990evaluating,justeson1995technical} and the methods at the time included the following two-step procedure: (1) extracting a list of candidate terms; and (2) determining which of these candidate terms are correct using either supervised or unsupervised techniques.  We briefly summarize different supervised ATE techniques according to their evolution below.

\subsection{Approaches based on term characteristics and statistics}
The first ATE approaches leveraged linguistic knowledge and distinctive linguistic aspects of terms to extract a possible candidate list. Several NLP techniques are employed to obtain the term's linguistic profile (e.g., tokenization, lemmatization, stemming, chunking, etc.). On the other hand, several studies proposed statistical approaches toward ATE, mostly relying on the assumption that a higher candidate term frequency in a domain-specific corpus (compared to the frequency in the general corpus) implies a higher likelihood that a candidate is an actual term. Some popular statistical measures include termhood \cite{vintar2010bilingual}, unithood \cite{daille1994towards} or C-value \cite{frantzi1998c}. Many current systems still apply their variations or rely on a hybrid approach combining linguistic and statistical information \cite{kessler2019extraction,repar2019termensembler}.

\vspace{-10pt}
\subsection{Approaches based on machine learning and deep learning}

The recent advances in word embeddings and deep neural networks have also influenced the field of term extraction. Several embeddings have been investigated for the task at hand, e.g., 
non-contextual \cite{zhang2018semre,amjadian2016local}, contextual \cite{kucza2018term} word embeddings, and the combination of both \cite{gao2019feature}. The use of language models for ATE tasks is first documented in the TermEval 2020 \cite{rigouts2020termeval} on the trilingual ACTER dataset. While the Dutch corpus winner used BiLSTM-based neural architecture with GloVe word embeddings, the English corpus winner \cite{hazem2020termeval} fed all possible extracted n-gram combinations into a BERT binary classifier. Several Transformer variations have also been investigated  \cite{hazem2020termeval} (e.g., BERT, RoBERTa, CamemBERT, etc.) but no systematic comparison of their performance has been conducted. Later, the HAMLET approach \cite{rigouts2021hamlet} proposed a hybrid adaptable machine learning system that combines linguistic and statistical clues to detect terms. Recently, sequence-labeling approaches became the most popular modeling option. They were first introduced by \cite{kucza2018term} and then employed by \cite{lang2021transforming} to compare several ATE methods (e.g., binary sequence classifier, sequence classifier, token classifier). Finally, cross-lingual sequence labeling proposed in \cite{conneau2020unsupervised,lang2021transforming,tran2022can} demonstrates the capability of multilingual models and the potential of cross-lingual learning. 

\vspace{-10pt}
\subsection{Approaches for Slovenian term extraction}
The ATE research for the less-resourced languages, especially Slovenian, is still hindered by the lack of gold standard corpora and the limited use of neural methods. Regarding the corpora, the recently compiled Slovenian KAS corpus \cite{erjavec2021kas}  was quickly followed by the domain-specific RSDO5 corpus \cite{jemec2021corpus}. Regarding the methodologies, techniques evolved from purely statistical \cite{vintar2010bilingual} to more machine learning based approaches. For example, \cite{ljubevsic2019kas} extracted the initial candidate terms using the CollTerm tool \cite{pinnis2019extracting}, a rule-based system employing a language-specific set of term patterns from the Slovenian SketchEngine module \cite{fivser2016terminology}. The derived candidate list was then filtered using a machine learning classifier with features representing statistical measures. Another recent approach \cite{repar2019termensembler} focused on the evolutionary algorithm for term extraction and alignment. Finally, \cite{tran2022transformer} was one of the first to explore the deep neural approaches for Slovenian term extraction, employing XLMRoBERTa in cross- and multilingual settings.

\section{Methods}
\label{sec:method}

We briefly describe our chosen datasets in Section \ref{subsec:dataset}, the general methodology in Section \ref{subsec:method} and the chosen evaluation metrics in Section \ref{subsec:evaluation_metric}.

\vspace{-10pt}
\subsection{Datasets}
\label{subsec:dataset}
The experiments have been conducted on two datasets: ACTER v1.5 \cite{rigouts2020termeval} and RSDO5 v1.1 \cite{RSDOcorpus11}. The ACTER dataset is a manually annotated collection of 12 corpora covering four domains, Corruption (corp), Dressage (equi), Wind energy (wind), and Heart failure (htfl), in three languages, English (en), French (fr), and Dutch (nl). It has two versions of gold standard annotations: one including both terms and named entities (NES), and the other containing only terms (ANN). Meanwhile, the RSDO5 corpus v1.1 \cite{RSDOcorpus11} includes texts in Slovenian (sl), a less-resourced Slavic language with rich morphology. Compiled during the RSDO national project, the corpus contains 12 documents covering four domains, Biomechanics (bim), Chemistry (kem), Veterinary (vet), and Linguistics (ling). 

\vspace{-5pt}
\subsection{Workflow}
\label{subsec:method}
We consider ATE as a sequence-labeling task \cite{tran2022can} with IOB labeling regime \cite{rigouts2021hamlet,lang2021transforming}. The model is first trained to predict a label for each token in the input text sequence and then applied to the unseen test data. From the token sequences labeled as terms, the final candidate term list for the test data is composed. 
\vspace{-10pt}

\subsubsection{Empirical evaluation of pretrained language models}
\begin{figure}[ht]
    \centering
    \includegraphics[scale=0.18]{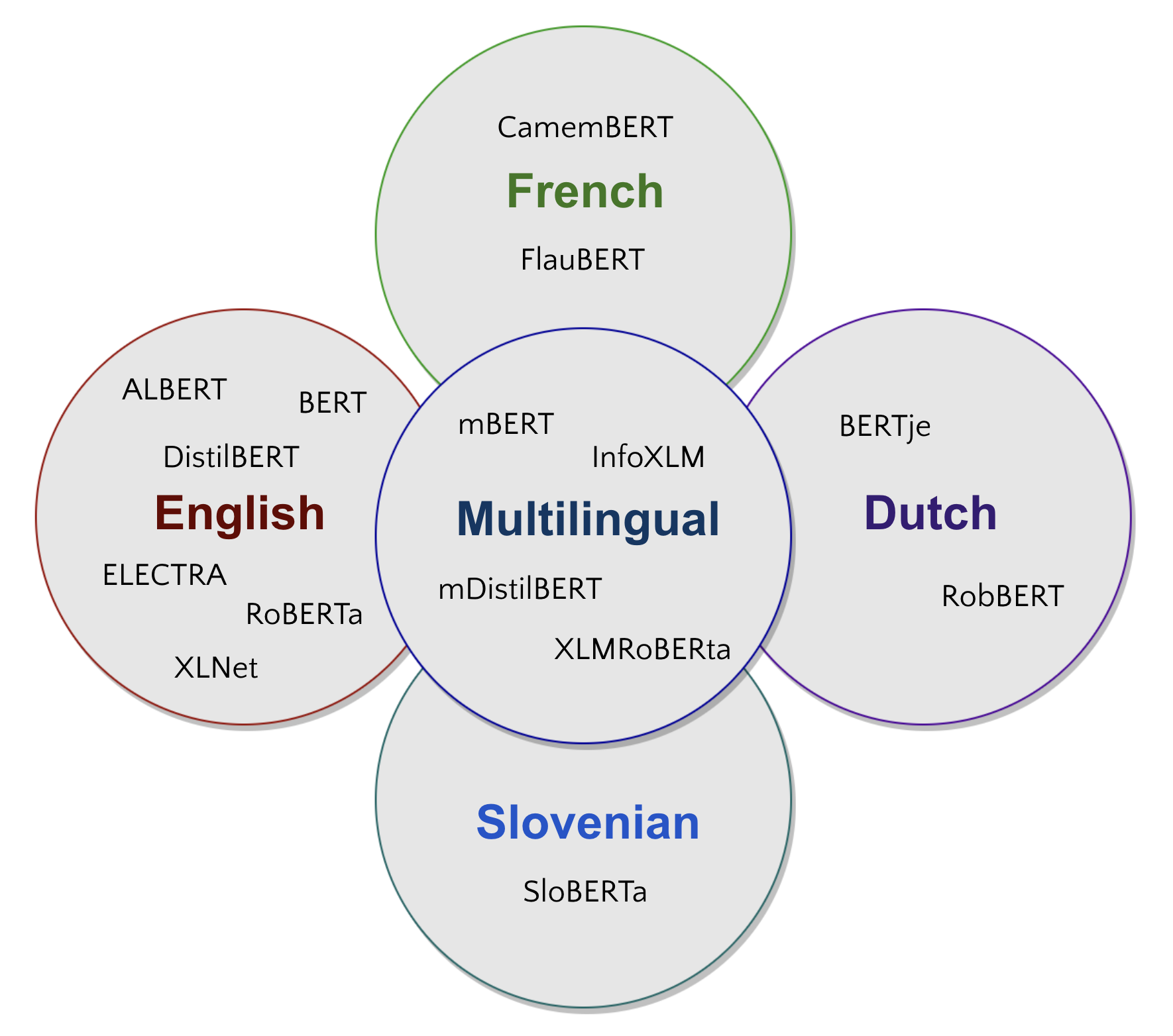}
    \caption{Empirical evaluation of pretrained language models on the ATE task.} 
    \label{fig:taxonomy}
\end{figure}

We conduct a systematic evaluation of mono- and multilingual Transformers-based models on the ATE task modeled as sequence labeling. The models were obtained from Huggingface\footnote{https://huggingface.co/models} according to the number of downloads and likes criteria. The chosen models are presented in Fig. \ref{fig:taxonomy}.
Regarding the multilingual systems, we investigate the performance of mBERT \cite{devlin2018bert} (\textit{bert-base-multilingual-uncased}), mDistilBERT \cite{sanh2019distilbert} (\textit{distilbert-base-multilingual-cased}), InforXLM \cite{chi2021infoxlm} (\textit{microsoft/ infoxlm-base}), and XLMRoBERTa \cite{conneau2019unsupervised} (\textit{xlm-roberta-base}). All the chosen multilingual models are fine-tuned in a monolingual fashion due to findings from the related work \cite{lang2021transforming,tran2022can} showing that no (or only marginal) gains are obtained if the model is fine-tuned on the multilingual training data.

Regarding the monolingual models, we evaluate several English autoencoding Transformer-based models, including ALBERT \cite{lan2019albert} (\textit{albert-base-v1} and \textit{albert-base-v2}), BERT \cite{devlin2018bert} (\textit{bert-base-uncased}), DistilBERT \cite{sanh2019distilbert} (\textit{distilbert-base-uncased}), ELECTRA (\textit{electra-small-generator}) and RoBERTa \cite{liu2019roberta} (\textit{xlm-roberta-base}), and one autoregressive model, XLNet \cite{yang2019xlnet} (\textit{xlnet-base-cased}). For French,  we use CamemBERT \cite{martin2019camembert} (\textit{camembert-base}) and FlauBERT \cite{le2020flaubert} (\textit{flaubert\_base\_uncased}), for Dutch, we employ BERTje (\textit{bert-base-dutch-cased}) and RobBERT (\textit{robBERT-base} and \textit{robbert-v2-dutch-base}) models, and for Slovenian, we choose SloBERTa (\textit{sloberta}), the RoBERTa-based model trained on a large Slovenian corpus.



\subsubsection{Ensemble of Transformer models}  
\begin{figure}[ht]
    \centering
    \includegraphics[scale=0.35]{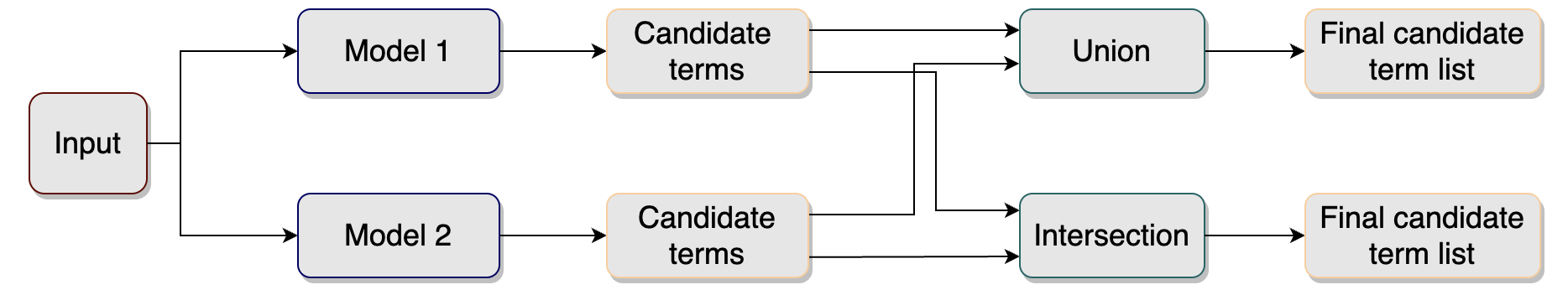}
    \caption{The general ensembling workflow.}
    \label{fig:workflow}
\end{figure}

Regarding results in Section 3.2.1, we propose a novel ensembling approach based on Transformer models for ATE task as we observe the general tendency for Precision to be better than Recall for all but few monolingual and multilingual models tested (see Tables \ref{tab:acter-monolingual} and \ref{tab:rsdo5_res}). This leads us to believe that by combing the outputs of different models, we could achieve improvements in Recall and by extension also in the overall F1-score. We consider two strategies for combining the outputs from different models of the ensemble, namely the union and the intersection of the candidate term lists from the models of the ensemble. See the entire procedure in Fig. \ref{fig:workflow}.
\vspace{-15pt}
\vspace{-15pt}

We hypothesize that by combining the outputs of two models, we might be able to significantly improve the Recall of the term extraction system. To validate this hypothesis, we test three combinations: Combine the outputs of the (1) best mono- and multilingual models; (2) two best monolingual models; and (3) two best multilingual models.


\vspace{-10pt}
\subsection{Evaluation metrics}
\label{subsec:evaluation_metric}
We evaluate each term extraction system by comparing the aggregated list of candidate terms extracted on the level of the whole test set with the manually annotated gold standard term list using Precision, Recall, and F1-score. These evaluation metrics have also been used in the related work \cite{hazem2020termeval,lang2021transforming,rigouts2020termeval}.
\vspace{-10pt}
\section{Results}
\label{sec:results}
\vspace{-5pt}
We first present the results of mono- and multilingual Transformer-based models obtained on ACTER and RSDO5 test sets compared with the SOTAs. Then, we demonstrate the impact of the ensemble post-processing step.
\vspace{-15pt}
\subsection{Monolingual evaluation}
\label{sec:mono_evaluation}

\subsubsection{ACTER corpus} Not many approaches have been tested on the ACTER corpus v1.5 due to its novelty. Thus, we apply the approach proposed by \cite{lang2021transforming} (i.e., employing XLMRoBERTa as a token classifier), which achieved SOTA on the previous corpus version, and consider it as a baseline. The Heart failure domain is used as a test set, the same as in TermEval 2020.
\vspace{-25pt}
\begin{table}[ht]
\caption{Results of monolingual term extraction on the ACTER dataset.}
\label{tab:acter-monolingual}
\begin{subtable}{1\linewidth}
\centering
    \scalebox{0.55}{
    \begin{tabular}{|c|c|ccc|ccc|} 
    \toprule
    \multirow{2}{*}{\textbf{ }} & \multirow{2}{*}{\textbf{Models }} & \multicolumn{3}{c|}{\textbf{ANN}}  & \multicolumn{3}{c|}{\textbf{NES}}  \\                   
                                      && \textbf{Precision} & \textbf{Recall} & \textbf{F1-score} & \textbf{Precision} & \textbf{Recall} & \textbf{F1-score} \\ 
    \midrule
    \multirow{4}{*}{Mono} & albert-base-v1	        & 52.58& 47.40& 49.86&	54.42& 54.63& 54.52\\
                                & albert-base-v2      	& 49.85& 48.50& 49.17&	57.01& 55.13& 56.05\\
                                & bert-base-uncased	    & \textbf{59.06}& 32.44& 41.88&	61.42& 47.50& 53.57\\
                                & distilbert-base-uncased & 58.24& 38.75& 46.54&	61.06& 48.24& 53.90\\
                                & electra-small-generator&56.46 & 46.80 & 51.18&58.17 & 47.31 & 52.18\\
                                & roberta-base      	& 58.10 & 51.04 & 54.34&\textbf{62.28} & 56.30 & \textbf{59.14}\\
                                & xlnet-base-cased	    & 56.50& \textbf{53.92}& \textbf{55.18}&	58.34& \textbf{57.30}& 57.82\\
                                
    \midrule
   \multirow{3}{*}{Multi} & bert-base-multilingual-uncased & 55.21& 35.24&  43.02& \textbf{62.06}& 49.44& 55.04\\
                                 & distilbert-base-multilingual-cased &55.14 & 45.45& 49.83&57.10& 54.20& 55.61\\
                                 & infoxlm-base & \textbf{57.67} & \textbf{54.64} & \textbf{56.11}&61.18 & \textbf{54.48} & \textbf{57.64}\\
                                 \rowcolor{blue!20}
                                 & xlm-roberta-base (baseline)    & 57.34& 51.46& 54.24&	58.80& 55.52& 57.11\\
    \bottomrule
    \end{tabular}
    }
\vspace{-5pt}
\caption{English corpus}
\vspace{-5pt}
    \scalebox{0.55}{
    \begin{tabular}{|c|c|ccc|ccc|} 
    \toprule
    \multirow{2}{*}{\textbf{ }} & \multirow{2}{*}{\textbf{Models }} & \multicolumn{3}{c|}{\textbf{ANN}}  & \multicolumn{3}{c|}{\textbf{NES}}  \\                   
                                      & & \textbf{Precision} & \textbf{Recall} & \textbf{F1-score} & \textbf{Precision} & \textbf{Recall} & \textbf{F1-score} \\ 
    \midrule
   \multirow{1}{*}{Mono} & camembert-base &	\textbf{70.51}& \textbf{44.97}& \textbf{54.92}	&\textbf{70.74}& \textbf{52.23}& \textbf{60.09}\\
                                 & flauberta  &75.91 & 26.17& 38.92	&75.28 & 39.01 & 51.39\\ 
    \midrule
   \multirow{2 }{*}{Multi}& bert-base-multilingual-uncased &	67.77& 37.66& 48.42	&69.39& \textbf{48.99}& 57.43\\
                                 & distilbert-base-multilingual-cased &64.45 &  43.45 & 51.91&65.20& 48.78& 55.81\\ 
                                 & infoxlm-base &68.74 & 39.77 & 50.39&\textbf{71.10} &48.90& \textbf{57.95}\\
                                 \rowcolor{blue!20}
                                 & xlm-roberta-base (baseline) &	\textbf{68.85}& \textbf{48.61}& \textbf{56.99}	& 70.71& 46.46& 56.08\\
    
    \bottomrule
    \end{tabular}}
\vspace{-5pt}
\caption{French corpus}
\vspace{-5pt}
    \scalebox{0.55}{
    \begin{tabular}{|c|c|ccc|ccc|} 
    \toprule
    \multirow{1}{*}{\textbf{ }} & \multirow{2}{*}{\textbf{Models }} & \multicolumn{3}{c|}{\textbf{ANN}}  & \multicolumn{3}{c|}{\textbf{NES}}  \\                   
                                      & & \textbf{Precision} & \textbf{Recall} & \textbf{F1-score} & \textbf{Precision} & \textbf{Recall} & \textbf{F1-score} \\ 
    \midrule
    \multirow{2 }{*}{Mono} & bert-base-dutch-cased &	65.59 & \textbf{65.53} & \textbf{65.56}& 67.61 & \textbf{66.02} & \textbf{66.81}\\
                                  & robBERT-base &  69.58& 36.84& 48.17 & 71.63 & 55.01 & 62.23\\
                                  & robbert-v2-dutch-base & \textbf{71.56} & 36.40 & 48.25	& \textbf{73.58} & 55.72 & 63.42\\
    \midrule
    \multirow{3}{*}{Multi}& bert-base-multilingual-uncased&	\textbf{70.67}& 62.49& 66.33	&72.34& 63.71& 67.75\\ 
                                & distilbert-base-multilingual-cased &69.80 & 61.28 & 65.26&69.45& \textbf{66.15}& 67.76\\
                                 & infoxlm-base & 70.43& 66.73& \textbf{68.53}&73.47& 64.24&\textbf{68.55}\\
                                 \rowcolor{blue!20}
                                 & xlm-roberta-base (baseline) &	68.53& \textbf{67.94}& 68.23	&\textbf{73.93}& 60.65& 66.63\\
    \bottomrule
    \end{tabular}}
\vspace{-5pt}
\caption{Dutch corpus}
\vspace{-5pt}
\end{subtable}%
\end{table}
\vspace{-15pt}

In general, multilingual pretrained models outperform the monolingual ones in Recall and F1-score when applied for extraction of the ANN annotations in all three languages. If named entities are included (NES), monolingual models outperform multilingual models in two (English and French) out of three languages in the ACTER dataset.  When it comes to individual models, InfoXLM outperforms other mono- and multilingual models in the F1-score on the Dutch corpus (for both ANN and NES) and on the English corpus (for ANN). If we compare the results of our study with the XLMRoBERTa baseline using the same monolingual settings from \cite{lang2021transforming}, our best-performing models surpass the baseline in all cases (e.g., the F1-score increases by 1.87\% on ANN and 1.5\% on NES in the English corpus;  4.01\% on French NES; 0.3\% on ANN and 1.92\% on NES in the Dutch corpus) except for the French ANN annotations.


\subsubsection{RSDO5 corpus} We also compare the performance of different mono- and multilingual models on the RSDO5 corpus, Here, we evaluate the models on all domains as demonstrated in Table \ref{tab:rsdo5_res}. By using two domains from the RSDO5 corpus for training, the third one for validation, and the last one for testing, all the models prove to have relatively consistent performance across different combinations. The monolingual SloBERTa model outperforms other approaches (including the XLMRoBERTa baseline from \cite{tran2022transformer}) in all cases by a relatively large margin in F1-score. By employing this model and looking at the best performing train/validation combinations for each test domain, we improve the SOTA baseline in the Linguistics domain by 2.21\%, in Veterinary by 2.35\%, in Chemistry by 5.26\%, and in Biomechanics by 2.66\% regarding F1-score. Our results, thus, set a new SOTA on the Slovenian corpus.

\vspace{-25pt}
\begin{table}
\caption{Results of monolingual term extraction on the RSDO5 dataset.}
\begin{subtable}{1\linewidth}
\centering
\scalebox{0.65}{
\begin{tabular}{|c|c|c|cccccc|cccccc|cccccc|} 
\toprule
\multirow{2}{*}{\textbf{Training}} & \multirow{2}{*}{\textbf{Val}} & \multirow{2}{*}{\textbf{Test}} & \multicolumn{6}{c|}{\textbf{xlm-roberta-base}} & \multicolumn{6}{c|}{\textbf{sloberta}}& \multicolumn{6}{c|}{\textbf{infoxlm-base}} \\
& &  && \textbf{Precion} && \textbf{Recall} && \textbf{F1-score} && \textbf{Precion} && \textbf{Recall} && \textbf{F1-score}  && \textbf{Precion} && \textbf{Recall} && \textbf{F1-score}           \\ 
\midrule
bim + kem & vet  & ling && 69.55 && 64.05 && 66.69 && 73.23 &&  70.51 && 71.84 && 68.37 && 71.38 && 69.84    \\
bim + vet & kem  & ling && 66.20 && 72.38 && 69.15 && 73.91 && 73.53 && 73.72&& 67.74&& 71.46&& 69.55     \\
kem + vet & bim  & ling && 69.48 && 73.66 && 71.51 && \textbf{74.45} && \textbf{73.96} &&  \textbf{74.20} &&  73.71&& 66.90 && 70.14 \\ 
\midrule
bim + kem  & ling & vet  && 71.06 && 66.72 && 68.82	&& 77.56 && \textbf{65.96} && \textbf{71.29} && 71.04&& 63.69&& 67.16  \\
bim + ling & kem  & vet  && 72.66 && 65.59 && 68.94	&& \textbf{78.33} && 65.31 && 71.23 && 66.88&& 68.93&& 67.89\\
ling + kem & bim  & vet   && 69.30 && 68.07 && 68.68	&& 76.66 && 64.89 && 70.29 && 72.69 && 63.63 && 67.86\\
\midrule
bim + vet  & ling & kem && 68.67  && 55.13 && 61.16  && 	72.14  && 65.88  && 68.87  &&  67.77&&60.40&& 63.87    \\
bim + ling & vet  & kem  && 70.23   && 59.24 && 64.27	  && 70.29  && \textbf{68.45}  && 69.36 && 72.00 && 56.58 && 63.37  \\
ling + vet & bim  & kem && 70.14  && 60.27 && 64.83  && 	\textbf{73.52}  && 66.96  && \textbf{70.09}  && 71.22 && 59.49&& 64.83    \\  
\midrule
vet + kem    & ling & bim && 	62.25 && 65.20 && 63.69	  && 	67.97 && 67.36 && 67.66&& 63.60&& 60.59&& 62.06\\
vet + ling   & kem  & bim && 	62.35 && 63.99 && 63.16	  && 	\textbf{68.97} && 66.62 && \textbf{67.77}	  &&  56.66&&  67.53 &&61.62 \\
ling + kem   & vet  & bim  && 	63.51 && 66.80 && 65.11	  && 	67.15 && \textbf{67.79} && 67.47	  &&  60.61&&64.04&& 62.28   \\ 
\bottomrule
\end{tabular}
}
\vspace{10pt}
\scalebox{0.65}{
\begin{tabular}{|c|c|c|cccccc|cccccc|} 
\toprule
\multirow{2}{*}{\textbf{Training}} & \multirow{2}{*}{\textbf{Val}} & \multirow{2}{*}{\textbf{Test}} & \multicolumn{6}{c|}{\textbf{bert-base-multilingual-uncased}} & \multicolumn{6}{c|}{\textbf{distilbert-base-multilingual-cased}}   \\
& &  && \textbf{Precion} && \textbf{Recall} && \textbf{F1-score} && \textbf{Precion} && \textbf{Recall} && \textbf{F1-score}         \\ 
\midrule
bim + kem & vet  & ling && 66.77&& 65.86  && 66.31  && 61.82 &&  53.38 && 57.29    \\
bim + vet & kem  & ling && 66.80 && 68.01&& 67.40 && 59.14  &&  67.20   &&  62.91 \\
kem + vet & bim  & ling && 65.97&& 69.62  && 67.75  &&  60.94  &&  58.16  &&  59.52 \\
\midrule
bim + kem  & ling & vet  &&   68.18&& 61.56&& 64.70  &&  63.76  &&  58.70   &&  61.13  \\
bim + ling & kem  & vet  &&   68.58 &&65.46&& 66.98 &&  65.83   &&  58.15   &&  61.75  \\
ling + kem & bim  & vet   &&  69.12 &&   60.61&& 64.59  && 66.01   &&  54.02  &&  59.42 \\
\midrule
bim + vet  & ling & kem &&  65.35 &&  59.73 && 62.41  && 55.73&& 60.52&& 58.03  \\
bim + ling & vet  & kem  &&  65.53  &&  63.22&& 64.35   && 60.15&& 55.83&& 57.91\\
ling + vet & bim  & kem &&   67.32 && 53.96 && 59.90  && 59.53&& 57.70 && 58.60  \\
\midrule
vet + kem    & ling & bim &&  62.63 &&  60.85 &&  61.73 && 57.84&& 55.84&& 56.82 \\
vet + ling   & kem  & bim &&   65.25  &&  58.30   &&  61.58 && 60.62&& 56.36&& 58.41\\
ling + kem   & vet  & bim  &&  62.69 &&   63.61&&  63.15 && 62.04 &&  52.44 &&  56.84 \\
\bottomrule
\end{tabular}
}

\end{subtable}%

\label{tab:rsdo5_res}
\end{table}

\vspace{-40pt}

\subsection{Transformer ensembling}
\begin{figure}[ht]
    \centering
    \includegraphics[scale=0.29]{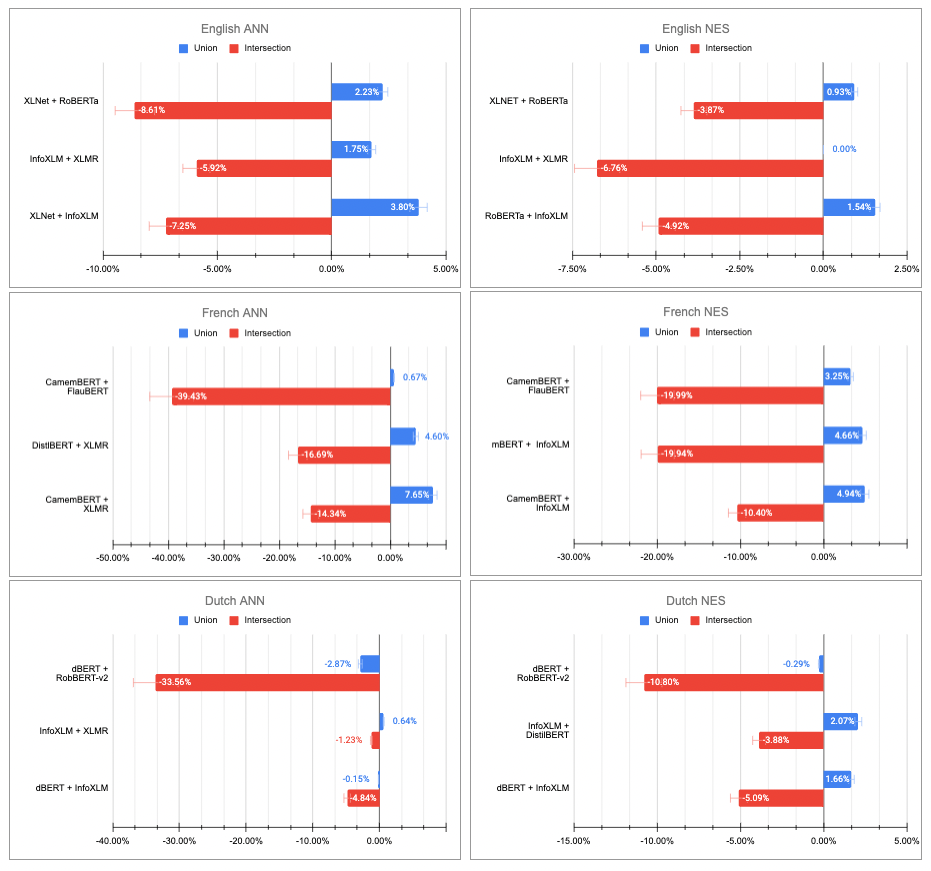}
    \caption{F1-score improvement by combining two best classifiers in ACTER.}
    \label{fig:improvement}
\end{figure}

We also evaluate the performance of the proposed ensembling approach described in Section 3.2.2. The improvements/decline in performance over the best single model on different languages of the ACTER dataset are shown in Fig. \ref{fig:improvement}. The results indicate that combining the acquired term sets of the two best-performing classifiers (no matter what type of classifiers they are) using the union always results in the biggest gain.

\vspace{-10pt}
\section{Conclusion}
\label{sec:conclusion}
\vspace{-10pt}
We proposed an empirical evaluation of different mono- and multilingual Transformers-based models on the monolingual sequence-labeling cross-domain term extraction. The experiments were conducted on the trilingual ACTER dataset and the Slovenian RSDO5 dataset. Furthermore, we tested how ensembling different mono- or multilingual models affects the performance of the overall term extractor. The results demonstrate that multilingual models outperform the monolingual ones in Recall and F1-score when applied for ANN extraction. Meanwhile, monolingual models capture the information about terms better than multilingual ones when it comes to the extraction of NES annotations. We also showed that by ensembling different Transformer models we can obtain further boosts in performance for all languages. As a consequence, we established the new SOTA on the ACTER and RSDO5 datasets.

In the future, we would like to take advantage of prompt engineering by considering ATE as a language model ranking problem in a sequence-to-sequence framework, where original sentences and statement templates filled by candidate terms are regarded as the source sequence and the target.

\noindent
\textbf{Acknowledgements} 
The work was partially supported by the Slovenian Research Agency (ARRS) core research programme Knowledge Technologies (P2-0103), and the Ministry of Culture of the Republic of Slovenia through the project Development of Slovene in Digital Environment (RSDO). The first author was partly funded by Region Nouvelle Aquitaine. This work has also been supported by the TERMITRAD (2020-2019-8510010) project funded by the Nouvelle-Aquitaine Region, France.

%
%
%

\bibliographystyle{splncs04}
\bibliography{icadl}

\end{document}